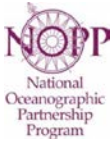 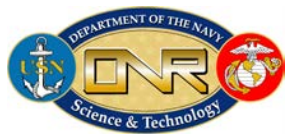 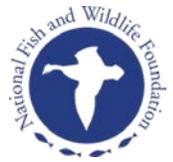

# DCL System Using Deep Learning Approaches for Land-based or Ship-based Real-Time Recognition and Localization of Marine Mammals

**Peter J. Dugan**
Bioacoustics Research Program,
Cornell Laboratory of Ornithology
Cornell University
159 Sapsucker Woods Road, Ithaca, NY 14850

phone: 607.254.1149    fax: 607.254.2460    email: pjd78@cornell.edu

**Christopher W. Clark**
Bioacoustics Research Program,
Cornell Laboratory of Ornithology
Cornell University
159 Sapsucker Woods Road, Ithaca, NY 14850

phone: 607.254.2408    fax: 607.254.2460    email: cwc2@cornell.edu

**Yann André LeCun**
Computer Science and Neural Science
The Courant Institute of Mathematical Sciences
New York University
715 Broadway, New York, NY 10003, USA

phone: 212.998.3283    mobile Phone: 732.503.9266    email: yann@cs.nyu.edu

**Sofie M. Van Parijs**
Northeast Fisheries Science Center, NOAA Fisheries
166 Water Street, Woods Hole, MA 02543

phone: 508.495.2119    fax: 508.495.2258    email: sofie.vanparijs@noaa.gov

## LONG-TERM GOALS

Overarching goals for this work aim to advance the state of the art for detection, classification and localization (DCL) in the field of bioacoustics. This goal is primarily achieved by building a generic framework for detection-classification (DC) using a fast, efficient and scalable architecture, demonstrating the capabilities of this system using on a variety of low-frequency mid-frequency cetacean sounds. Two primary goals are to develop transferable technologies for detection and classification in, *one*: the area of advanced algorithms, such as deep learning and other methods; and *two*: advanced systems, capable of real-time and archival processing. For each key area, we will focus on producing publications from this work and providing tools and software to the community where/when possible. Currently massive amounts of acoustic data are being collected by various institutions, corporations and national defense agencies. The long-term goal is to provide technical capability to analyze the data using automatic algorithms for (DC) based on machine intelligence. The goal of the automation is to provide effective and efficient mechanisms by which to process large acoustic datasets for understanding the bioacoustic behaviors of marine mammals. This capability will provide insights into the potential ecological impacts and influences of anthropogenic ocean sounds. This work focuses on building technologies using a maturity model based on DARPA 6.1 and 6.2 processes, for basic and applied research, respectively.

## OBJECTIVES

For each key area in the long-term goals, there are five basic objectives for this work: (1) develop a system to incorporate advanced acoustic DC algorithms using high performance computing (HPC), (2) develop advanced DC algorithms which can handle a variety of cetaceans, (3) create a framework for analyzing large amounts of data, using multiple DC algorithms, (4) collaborate on existing projects and with external parties and (5) transition technologies to the wider community, such as other groups and government programs. During project Phase I (December 2011 – September 2012) the team focused on the first objective. We developed software and hardware technologies capable of running in parallel or distributed systems. The software, called DeLMA, is capable of running on laptops or large cluster computers. For small applications, a standard workstation or laptop can achieve 12x throughput, in comparison to serial. Medium and large scale processing occurs on a specially designed hardware system, built during this research, called the acoustic data accelerator, or HPC-ADA machine. For our work, the ADA machine is designed to handle 64x in throughput, but scalable to sizes greater than 128x. For algorithms, a new structure for identifying sound objects called the Acoustic Segmentation Recognition (ASR) technology and capable of handling various species for detection-classification, including frequency modulated and pulse train sound patterns. These technologies are built on high performance computing, allowing for advanced algorithms to "plug-n-play" with small- and large-scale computer systems. Phase II of the project (October 2012 - September 2013), focused on (1) further developing advanced DC algorithms, (2) applying these algorithms to tackle "big" data problems in bioacoustics (3) collaborating on existing projects-publications and (4) transitioning technologies to a wider community by promoting the publication of open source solutions through data competitions through International Conference on Machine Learning (ICML).

## APPROACH

This research project is developing advanced methods for processing and exploring passive acoustic data, specifically new approaches for DC (deep learning) and advanced technology (HPC). To facilitate this work, Cornell has assembled an Integrated Research Team (IRT) of scientists, biologists and engineers. Members of this work are highly qualified research professionals, with experience ranging from acoustical engineering, signal processing and machine learning to biology. Coordinating the efforts for this work are Dr. Peter Dugan (PI) and Dr. Christopher Clark (co-PI). Collaborators include Dr. Yann LeCun (co-PI), New York University (NYU) and Dr. Sofie Van Parijs (co-PI), Northeast Fisheries Science Center (NEFSC), Woods Hole. Specialized talents for this research are broken down into three main groups. Dr. Peter Dugan leads the Machine Learning Systems Integration Team at Cornell University. This group leverages experience in applied recognition systems and focus development and integration tasks on advanced technologies for DC. Dr. Dugan leverage four senior consultants, Dr. Harold Lewis, Dr. Mark Fowler, Katie Vannicola and Dr. Rosemary Paradis; with expertise as senior professionals and faculty in the areas of computational intelligence, signal processing, speech processing and neural networks, respectively. Dr. John Zollweg, Marian Popescu, Dr. Yu Shiu, Mohammad Pourhomayoun, Katie Vanicolla and Adam Mikolajczyk form a sub-group focused on HPC technology and software development. Dr. Yann LeCun's group at NYU, is focused on basic research and development for applying deep learning technologies to detect and classify underwater sounds in real-time. The Biology Team facilitates data analysis and biological direction; Dr. Christopher Clark provides expertise and leadership in marine mammal bioacoustics, along with Dr. Sofie Van Parijs. Together, Cornell, NYU and NEFSC are undertaking basic and applied research, while coordinating publication efforts in engineering, biology, and systems operational research.

**Work Completed**

From October 2012 through September 2013 our research focused on five major initiatives:
1. *International workshops, conferences and data challenges*
2. *Enhancements of the ASR algorithm for frequency-modulated sounds: Right Whale Study*
3. *Enhancements of the ASR algorithm for pulse trains: Minke Whale Study*
4. *Mining Big Data Sound Archives using High Performance Computing software and hardware*
5. *Large Pulse Train Study: Minke Vocal Activity East Coast United States*

1. *International workshops, conferences and data challenges*
Workshops and conferences, significant focus was devoted to two international events, DCLDE (Detection, Classification, Localization and Density Estimation) and ICML-WMLB (International Conference on Machine Learning; Workshop on Machine Learning for Bioacoustics) [1]. The DCLDE assembled various leading experts focused on terrestrial and marine applications. Cornell provided two posters [2, 3] and a talk on new approaches for North Atlantic right whale (*Eubalaena glacialis*) detection methods [4]. The ICML-WMLB was hosted in joint support through a series of collaborators connected through the ONR grant [1]. Cornell provided two key-note talks, one focusing on big data considerations for bioacoustics [6] and the second on marine mammal sounds [7]. Three other papers were also delivered, which summarize significant works on minke whale (*Balaenoptera acutorostrata*) and right whale signal processing as they relate to big data processing applications published through this grant [5, 8, 9].

We participated in a total of three DC competitions. First challenge was hosted by DCLDE and focused on creating detection algorithms for data mining for right whale up-calls. DCLDE challenge used continuous sounds recorded during a prior NOPP grant. There were approximately 8 entries that competed (results not published by DCLDE), and Cornell submitted 4 of the 8 unique entries for up-call signatures. There were three categories for scoring, Cornell finished first in two of three categories using algorithms published in the following material [5, 6], these methods are referred to as ASR-FM$_{HOG}$ and ASR-FM$_{CRA}$.

The second and third data challenges were offered through Kaggle, promoting international participation amongst machine learning community. The first Kaggle challenge, called Kaggle(1) for

*Dugan, Clark, LeCun and Parijs*

this report, was sponsored by Cornell and Marine Explore; data was taken from Cornell's Auto-buoy system [12]. The second challenge through Kaggle, referred to as Kaggle(2) for this report, was sponsored through ICML-WMLB conference. The problem outlined for Kaggle(1) was identical to Kaggle(2), except sounds came from two different sources, auto-buoy and marine autonomous recorder (MARU), respectively. Both Kaggle competitions focused on designing the optimal classifiers for right whale up-call sounds. Together both competitions had over 200 unique international entries.

With respects to all three competitions, several lessons were derived from this work and are summarized in [10, 11]. DCLDE used the entire sound stream, whereby teams were required to provide a detector and classifier algorithm. Kaggle(1) and Kaggle(2) competition focused on sound clips, not continuous sounds allowing the calls to be isolated within a narrow window of time, eliminating the need to perform extensive detection stages; allowing teams to focus on classification only. For Kaggle(1) and Kaggle(2), the top entry was a hand designed algorithm that contained several stages similar to those found in convolutional neural networks. Top entries also included deep learning algorithms from several competitors. The top solutions required retraining the classification algorithms based on Auto-buoy to MARU data. Cornell results show as much as a 13% difference between sources when using CRA and HOG algorithms (Figure 3). Lastly, the Cornell team measured a high degree of variation in human labeled data that had been validated by analysts. Analysts at NOAA's North East Fisheries Science Center (NEFSC) and Cornell worked independently to provide labels for the DCLDE data challenge: NEFSC supplied truth for DCLDE, and Cornell used the same set for the ICML contest, see Table 1. Both groups used four days of data extracted from Cornell's NOPP 2007-2010 grant (Table 1).

2. *Enhancements of the ASR-FM algorithm technology: Right Whale Data Challenge*

Two basic types of ASR algorithms were developed to address a variety of mysticete sounds, these include frequency modulated (ASR-FM) and pulse train (ASR-PT) sounds. ASR-FM algorithms detect short duration frequency-modulated sounds (e.g. right whale up-call, humpback whale, *Megaptera novaeangliae*, sounds), while ASR-PT algorithm detects repeating sounds (e.g. fin whale *Balaenoptera physalus*, minke whale *Balaenoptera acutorostrata*, humpback whale). The research investigated several variations of DC technologies. ASR algorithm work showed several promising components. First the Maximally Stable Extremal Regions (MSER) process was applied to detect on the spectrogram regions of interest (ROI) which possess distinguishing, invariant and stable property [13] provide to be the most effective detection mechanism tested for a wide range of right whale up calls. For classification, final results yielded two main approaches, connected region analysis (CRA) and histogram of oriented gradients (HOG); these two types are referred to herein simply as ASR-$FM_{CRA}$ and ASR-$FM_{HOG}$. For ASR-$FM_{CRA}$, we develop a novel method based on machine-learning and image processing to identify North Atlantic right whale (NARW) up-calls in the presence of high levels of ambient and interfering noise. We apply a continuous region algorithm on the spectrogram to extract the regions of interest, and then use grid masking techniques to generate a small feature set that is then used in an artificial neural network classifier to identify the NARW up-calls. It is shown that the proposed technique is effective in detecting and capturing even very faint up-calls, in the presence of ambient and interfering noises, Figure 1. The method is evaluated on a dataset recorded in Massachusetts Bay, United States and results are published in [5]. The second computer vision technique, ASR-$FM_{HOG}$, uses a feature representation calculated through Histogram of Oriented Gradient (HOG) [14]. HOG not only captures the edge or gradient structure, see Figure 2, which is an important characteristics of shape, but also is robust to translations or rotations within a controllable degree of invariance [14]. Short-Time Fourier Transform (STFT) is used to generate the spectrogram from the feature vector received from the MSER Detection stage; a modified power-law algorithm is

*Dugan, Clark, LeCun and Parijs*

used for "de-noising" [15]. Regions of interest are determined from the MSER values along with a HOG feature extraction stage. Ada-boost is deployed for the classification between calls and noise (no-calls). Results for the ASR algorithm work, both ASR-FM$_{CRA}$ and ASR-FM$_{HOG}$ were configured to compete in the right whale data challenges for the DCLDE workshop [16] and the two international competitions hosted by Kaggle. [10, 11]. The Cornell Team competed both ASR-FM$_{HOG}$ and ASR-FM$_{CRA}$ and finished first in two of three categories for DCLDE and in the top 20% for both Kaggle competitions.

*3. Enhancements of the ASR-PT algorithm technology: Minke Whale Study*
The ASR-PT work outlines an approach for automatic DC of periodic pulse train signals using a multi-stage process based on spectrogram edge detection, energy projection and classification. Through this research this method has been successfully implemented to automatically detect and recognize pulse trains from minke whales (songs) and sperm whales (*Physeter macrocephalus*, foraging click trains), Table 2. A long-term goal of this work is to develop an algorithm that can be used on a variety of animals such as the blue whale (*Balaenoptera musculus sp.*) and fin whale (*Balaenoptera physalus*). Several focused research projects were conducted in collaboration with NEFSC, NYU and Cornell to identify and detect minke songs, along with a majority of the sounds within a large collection of passive acoustic data collected off of Massachusetts during a previous ONR-NOPP project (e.g. Stellwagen Bank National Marine Sanctuary, SBNMS) and other U. S. east coast locations. The detection methodology has been evaluated using 232 continuous hours of acoustic data, and a qualitative analysis of machine learning classifiers and their performance were published in [9, 17]. The ASR-PT algorithm was further enhanced from Phase I and applied to a large library of minke sounds spanning the U. S. east coast. This effort, conducted by co-PI's, achieved good processing results [17]. Prior to this work, few technology papers have been published on this topic [6, 8, 9]. The trained automatic detection and classification system was applied to 120 continuous hours, comprised of various challenges such as broadband and narrowband noises, low SNR, and other pulse train signature types. This automatic system achieves a TPR of 63% for FPR of 0.6% (or 0.87 FP/h), at a Precision (PPV) of 84% and an F1 score of 71%; results published [9].

*4. Mining Big Data Sound Archives using High Performance Computing software and hardware*
Scalable, high performance computing (HPC) is a corner stone of our work, with the goals of providing efficient computing strategies, lower error rates, wider community use and faster processing. Software from this grant has been applied on a variety of data sources, including Cornell Marine Autonomous Recording Unit (MARU), JASCO remote recorders and cabled data. The focus of our detection work consists of studying time series data and providing algorithms to properly classify acoustic objects that occur in a variety of ocean regions. Phase II of this effort has focused on processing vast amounts of acoustic data collected from a variety of sensors in various U.S. regions, including: Northeast (SBNMS), Mid-Atlantic (VA), Gulf of Mexico (GoMex), and Florida (NAVFAC), see Table 2 for a list of projects.

*5. Large Pulse Train Study: Minke Vocal Activity East Coast United States*
Based on results from the minke whale project, several questions motivated the team to explore to create new tools and interface existing ones. Research indicated that the design of the DC algorithms involve a complex mixture of advanced programming, which required new methods for visualizing and working with sound data. To incorporate this work, data science requires a workbench of visual aids to properly construct the algorithms. Large data also makes it difficult for human operators to explore the information. Many times humans need to sift through data events that are on the order of x$10^6$ and higher. Merging the DC algorithms with advanced visualization tools often provides

*Dugan, Clark, LeCun and Parijs*

unexpected insights and enables an expert human knowledge to "see" results and envision further solutions in entirely novel ways. One major challenge with running automatic DC algorithms on big data is validation. Often researchers have insufficient training data in comparison to the large archive. This means higher error rates and miss-classifications on the larger datasets. We conducted a study in an attempt to investigate whether or not human-expert-knowledge could help reduce detection-classification errors in large DC systems. The test utilized human experts by having them score sounds based on the signal quality. The human results and machine features were combined, using an artificial neural net, to automatically determine finalized score values. The automatic algorithm was applied to the entire dataset, minke sound events were filtered based on score. Results of the process we implemented, and published in [8], which shows a significant improvement in DC system performance, yielding as much as a 20% improvement in true positive rate for a given false positive rate when processing large datasets.

**RESULTS**

Workshops, conferences and competitions: Cornell submitted 4 of the 8 unique entries to the DCLDE Workshop and finished first in two of three categories for up-call detection using ASR-$FM_{HOG}$ and ASR-$FM_{CRA}$ methods. For the ICML workshop, this grant helped fund the data challenge along with several keynote speakers and papers specifically related to applied detection and classification of cetacean sounds. The data challenge attracted over 240 unique entries to compete for the best algorithm to classify right whale up-calls. Cornell used the same algorithms from the DCLDE (ASR-$FM_{HOG}$ and ASR-$FM_{CRA}$) and finished in the top 20% of the entries. Many of the top entries for Kaggle used algorithms with similar structures to Cornell's solution; i.e., template stage, feature extraction stage and a classifier stage. Cornell found that the classification stage relies heavily on the detection stage for ASR system. For example, small changes in one stage causes sensitivity errors in the other, this is referred to as shift invariance. Therefore, a DC model that is trained together might be a more efficient solution than separate algorithms chained together. Some of the top solutions were convolutional neural networks along with heuristic algorithms that also had similar processing models to Cornell's. In order to meet the top performance requirements for the final entries, all algorithms had to be retrained between both Kaggle competitions; where the first competition used Auto-buoy data, and the second used MARU data. Cornell found as much as a 13% difference between training performance when using ASR-$FM_{HOG}$ and ASR-$FM_{CRA}$ algorithms. It was also discovered that ground truth error rates far exceeded the error rates from the automatic classification. Two independent groups of observers hand validated right whale up-call events, yielding as much as a 23% difference between their DC results. Efforts from this grant developed various significant papers, manuscripts and presentations that focused on advanced detection and classification applied to high performance computing. The team provided 2 key-note talks at the ICML, 4 manuscripts at refereed conferences, 1 journal paper and several posters and non-paper talks. Collectively, the tools created through this grant have processed over 1,000,000 channel hours of acoustic data, and have contributed to a host of projects listed in Table 2. The ASR technology has been used to develop DC algorithms for processing the sounds from bryde's, fin, minke, right and sperm whales, as well as DC algorithms to find anthropogenic sounds such as seismic airgun sounds. Experiments were conducted to explore whether or not human experts could aide in browsing large datasets. This work improved sound classification performance by combining signal features, derived from the time-frequency spectrogram, with human perception based on 24 months of nearly continuous recordings [8]. The method exploits an artificial neural network (ANN) and learns the signal features based on the human-expert-knowledge. Lastly, the work for phase II was published by MathWorks as a case study for successfully using tools to deploy high performance computing solutions [18].

*Dugan, Clark, LeCun and Parijs*

# Figures

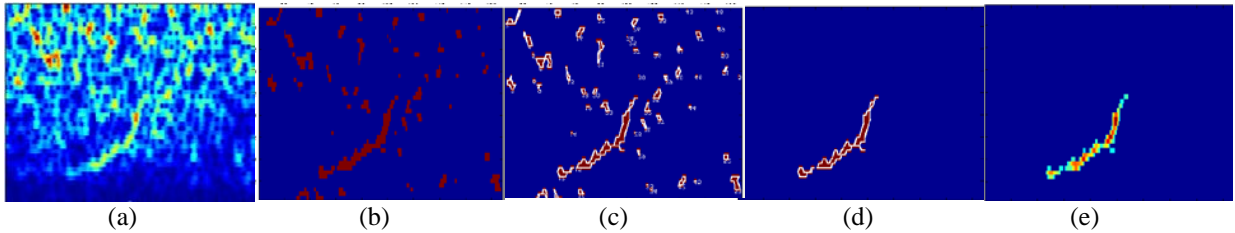

(a) (b) (c) (d) (e)

Figure 1. ASR-FM$_{CRA}$, Continuous Region Processing: (a) original Spectrogram; (b) spectrogram after denoising, normalization, equalization and binarization; (c) continuous region detection; (d) detected region of interest; and (e) the algorithm's output.

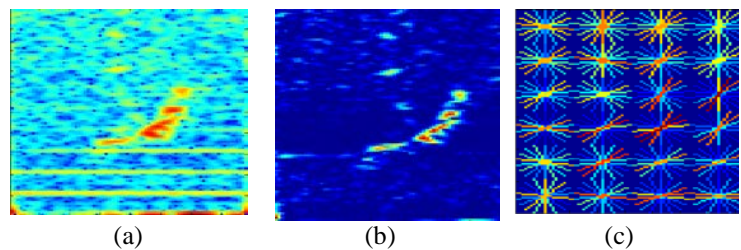

(a) (b) (c)

Figure 2. ASR-FM$_{HOG}$, Power-law de-noising: (a) the original spectrogram; (b) the spectrogram after power-law is applied; (c) HOG feature representation.

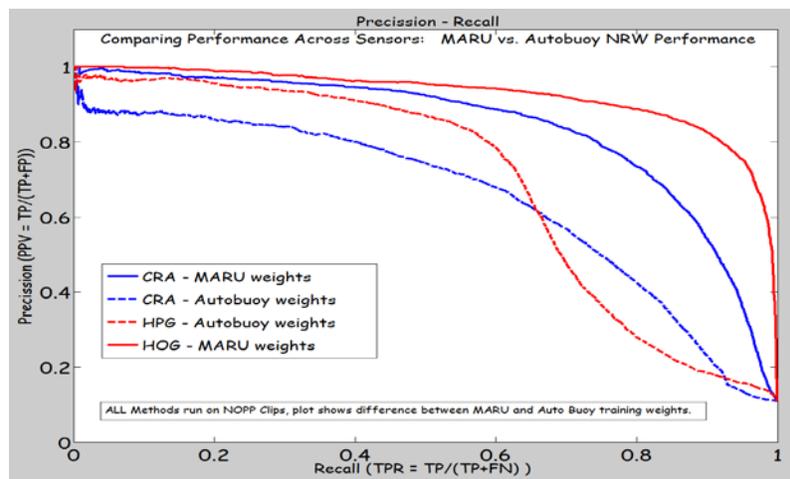

Figure 3. Precision-recall variation between sensor types (solid) shows the MARU sensor, (dotted) shows the Auto-buoy sensor. Right whale algorithms were run on both types of data: (blue) CRA based algorithm, (red) HOG based algorithm. Algorithms were first trained on the MARU data and then applied (without training) to the Auto-buoy data. Kaggle(1) and Kaggle(2) competitions used Auto-buoy and MARU data, respectively.

*Dugan, Clark, LeCun and Parijs*

| Day | Number of Calls Group A | Number Calls Group B | % Difference |
|---|---|---|---|
| 3-28-2009 | 767 | 656 | 15.6 |
| 3-29-2009 | 2280 | 1611 | 34.4 |
| 3-30-2009 | 1663 | 1265 | 27.2 |
| 3-31-2009 | 2206 | 1745 | 23.3 |

Table 1.  Results summarizing the differences between two independent groups of human operators that provided ground truth for right whale data challenges for DCLDE Workshop; Group A (NEFSC), Group B (Cornell-BRP).

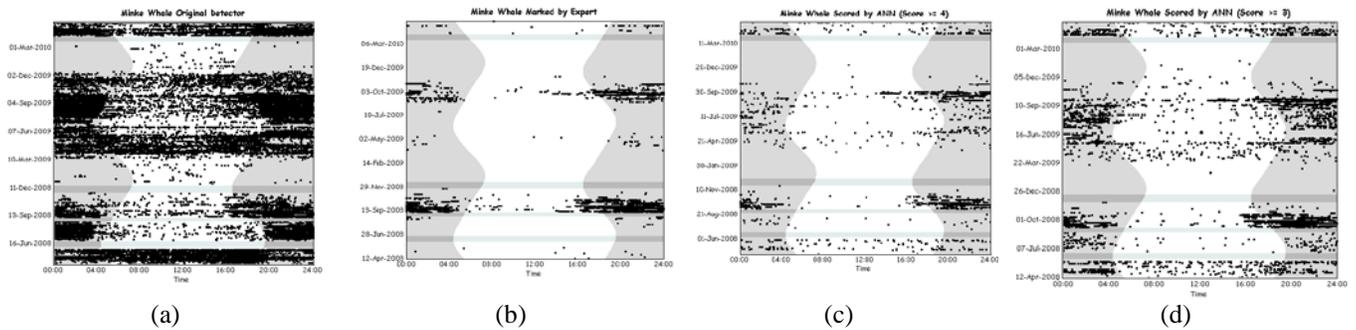

(a) (b) (c) (d)

Figure 4. Date versus time diel [19] patterns for test dataset on minke whale. (a) Original detection-classification by existing algorithm. (b) True detections by human expert. (c) Detection by ANN with score = 4. (d) Detection by ANN with score > 3

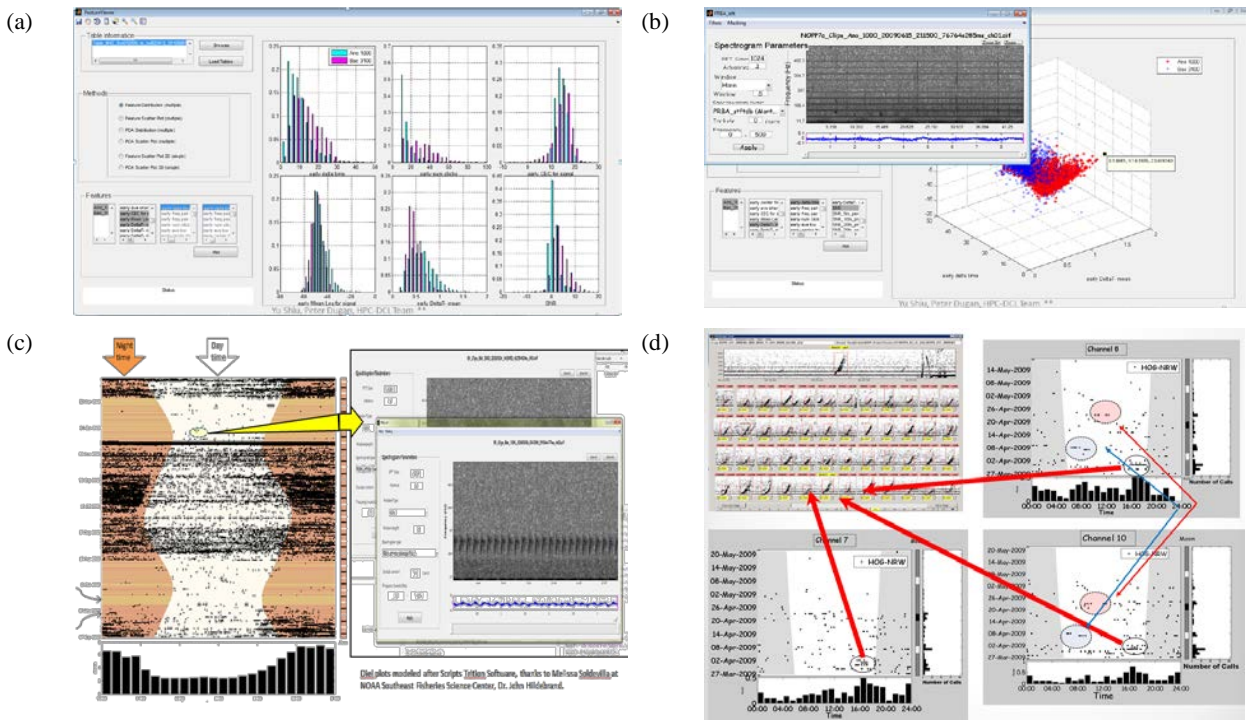

Figure 5.  HPC software applications on 24 month dataset: (a) classifier feature distribution and correlation, (b) object spectrogram identification from feature space distribution, (c) diel visualization with spectrogram browsing capability, (d) diel interoperability with event viewer.

*Dugan, Clark, LeCun and Parijs*

# IMPACT/APPLICATIONS

Currently the authors do not know of anyone who has successfully integrated HPC technology for doing advanced detection-classification for marine mammals, processing over $x10^6$ channel hours of sounds to date. This software has been applied to process passive acoustic archival data using multiple configurations developed during Phase I. The present software can now serve as a starting point for various applied reseach and development environments, such as at Naval Ocean Processing Facility (NOPF) or other university centers that wish to host HPC technologies for passive acoustic research.

# RELATED PROJECTS

During Phase II the HPC software system was applied to and supported the following projects.

| Deployment | Est. Channel Hours | ASR - Frequency Modulated | | | ASR - Pulse Train | | | |
| --- | --- | --- | --- | --- | --- | --- | --- | --- |
| | | Right Whale | Black Drum | Bryde's Whale | Seismic Exploration | Minke Whale | Fin Whale | Sperm Whale |
| Excellerate | 832k | X | - | - | - | - | - | - |
| GoMex | 350k | - | - | X | - | - | - | X |
| CAIRN Energy | 7k | - | - | - | X | - | - | - |
| Mass Clean Energy | 25k | X | - | - | - | X | X | - |
| Gulf of Maine | 26k | X | - | - | - | X | X | - |
| Cape Cod Bay 2008 Spring | 22k | X | - | - | - | X | X | - |
| Cape Cod Bay 2009 Spring | 22k | X | - | - | - | X | X | - |
| Cape Cod Bay 2010 Spring | 22k | X | - | - | - | X | X | - |
| Cape Cod Bay 2011 Spring | 22k | X | - | - | - | X | X | - |
| Cape Cod Bay 2012 Spring | 22k | X | - | - | - | X | X | - |
| SBNMS 08 Fall | 19k | X | - | - | - | X | X | - |
| SBNMS 08 Winter | 19k | X | - | - | - | X | X | - |
| SBNMS 09 Spring | 15k | X | - | - | - | X | X | - |
| SBNMS 08 Summer | 8k | X | - | - | - | X | X | - |
| New Jersey (pre 2011) | Data staged | -- | - | - | - | - | - | - |
| New York Long Island (pre 2011) | Data staged | -- | | - | - | - | - | - |
| VA 2012 Spring | 3k | X | X | - | - | X | x | - |
| VA 2012 Summer | 13k | X | X | - | - | X | X | - |
| VA 2012 Fall | 7k | X | X | - | - | X | X | - |
| VA 2012 Winter | ~7k | X | X | - | - | X | X | - |
| NAVFAC-01 (32 kHz) | 4k | X | - | - | - | X | X | X |
| NAVFAC-01 (32 kHz) | 6k | X | - | - | - | X | X | X |

Table 2. Projects that used the DeLMA HPC software system. List contains the major projects with estimated channel hours processed through this project's research tools; signal types described as frequency-modulated or pulse train events, which are derived from the acoustic-segmentation-recognition algorithm architecture.

# PUBLICATIONS

Dugan, P. J., and Clark, C. W., "Cornell Bioacoustics Scientists Develop a High-Performance Computing Platform for Analyzing Big Data", *MathWorks Central, User Stories*, http://www.mathworks.com/company/user_stories/Cornell-Bioacoustics-Scientists-Develop-a-High-Performance-Computing-Platform-for-Analyzing-Big-Data.html, October 2013. [published, refereed web article]

Dugan, P.J., M. Pourhomayoun, Y. Shiu, R. Paradis, A. Rice and C. Clark, Using High Performance Computing to Explore Large Complex Bioacoustic Soundscapes: Case Study for Right Whale Acoustics, Complex Adaptive Systems, Elsevier, Science Direct, Baltimore, MD, 2013. [published, refereed]

Clark, C.W., P.J. Dugan, Y. Le Cun, S. Van Parijs, D. Ponirakis and A. Rice, Application of advanced analytics and high-performance-computing technologies for mapping occurrences of acoustically active marine mammals over ecologically meaningful scales, *Key Note Talk*, ICML 2013, Workshop on Machine Learning for Bioacoustics, Atlanta, Georgia; USA. [published, keynote talk]

Pourhomayoun, M., P. Dugan, M. Popescu, D. Risch, H. Lewis and C. Clark, Classification for Big Dataset of Bioacoustic Signals Based on Human Scoring System and Artificial Neural Network, *ICML 2013 Workshop on Machine Learning for Bioacoustics*, Atlanta, GA, 2013. [published, refereed]

Dugan, P.J., M. Popescu, D. Risch, J. Zollweg, A. Mikolajczyk and C. Clark, Exploring Passive Acoustic Data Using High Performance Computing, Case Study for Pulse Train Exploration: Stellwagen Bank National Marine Sanctuary, *International Workshop on Detection, Classification, Localization and Density Estimation* (DCLDE), St. Andrews, Scotland, June 2013. [published, poster]

Popescu, M., P. Dugan, J. Zollweg, A. Mikolajczyk and C. Clark, Large-scale Detection and Classification (DC): Four Case Studies Using an Applied Distributed High Performance Computing (HPC) Platform, *International Workshop on Detection, Classification, Localization and Density Estimation* (DCLDE), *workshop poster*, St. Andrews, Scotland, June 2013. [published, poster]

Dugan, P.J., Y. LeCun, S. Van Parijs, D. Ponirakis, M. Popescu, M. Pourhomayoun, Y. Shiu, A. Rice and C. Clark, HPC and Bioacoustics, Practical Considerations for Detection Classification for Big Data, *Key Note Talk*, ICML 2013, *Workshop on Machine Learning for Bioacoustics*, Atlanta, Georgia; USA. [published, keynote talk]

Pourhomayoun, M., P.J. Dugan, M. Popescu and C. Clark, Bioacoustic Signal Classification Based on Continuous Region Processing, Grid Masking and Artificial Neural Network, *ICML 2013 Workshop on Machine Learning for Bioacoustics*, Atlanta, GA, 2013. [published, refereed]

Dugan, P. J., M. Pourhomayoun, Y. Shiu, M. Popescu, I. Urazghildiiev, X. Halkias, Y. LeCun and C. Clark, Survey of methods: Comparison of Automated Recognition for Right Whale Contact Calls, *International Workshop on Detection, Classification, Localization and Density Estimation* (DCLDE), St. Andrews, Scotland, June 2013. [published, refereed]

Dugan, P. J., W. Cukierski, Y. Shiu, A. Rahaman and C. Clark, "Kaggle Competition, Cornell University, The ICML 2013 Whale Challenge - Right Whale Redux, Kaggle.com, June 17, 2013,". [published, data challenge, refereed]

Risch, D., C. W. Clark, Dugan, P. J., Popescu, M., Siebert, U., and Van Parijs, S. M. 2013. Minke whale acoustics behavior and multi-year seasonal and diel vocalization patterns in Massachusetts Bay, USA. Mar. Ecol. Progr. Ser. 489: 279-295. [published, peer reviewed]

Popescu, M., P.J. Dugan, M. Pourhomayoun, D. Risch, H. Lewis and C. Clark, Bioacoustical Periodic Pulse Train Signal Detection and Classification using Spectrogram Intensity Binarization and Energy Projection, *ICML 2013 Workshop on Machine Learning for Bioacoustics*, Atlanta, GA, 2013. [published, refereed]

# PATENTS

Dugan, P, J., Clark, C.W., Ponirakis, D.W., Pitzrick, Rice, A., M.S., Zollweg, J.A., "System and Methods of Acoustic Monitoring", Provisional Patent Application, Case No. 5643-01-US, September 29, 2012.